# Authorship Verification – An Approach based on Random Forest
## Notebook for PAN at CLEF 2015

Promita Maitra, Souvick Ghosh and Dipankar Das

Department of Computer Science & Engineering, Jadavpur University, India
{ promita.maitra, souvikgiway, dipankar.dipnil2005 }@gmail.com

**Abstract.** Authorship attribution, being an important problem in many areas including information retrieval, computational linguistics, law and journalism etc., has been identified as a subject of increasingly research interest in the recent years. In case of Author Identification task in PAN at CLEF 2015, the main focus was given on cross-genre and cross-topic author verification tasks. We have used several word-based and style-based features to identify the differences between the known and unknown problems of one given set and label the unknown ones accordingly using a Random Forest based classifier.

## 1  Introduction

Author identification has a long history that includes some famous disputed authorship cases as well as various forensic applications. The task is used in order to determine the author who wrote a chapter or passage of a book (e.g., the holy Bible, being the most famous example). In general, the researches on author identification make use of the structure of the text and the words used to describe the contexts. In stylometry research, it is generally accepted that the authors have unconscious writing habits that can be evident from their use of words and grammar or punctuation etc. as such indicators could be reliable to identify an author [1][2][3][4]. The individual differences in use of language are referred to as idiolect (our idiolect includes the vocabulary appropriate to our various interests and activities, pronunciations reflective of the region in which we live or have lived, and variable styles of speaking that shift subtly depending on whom we are addressing.). The same feature, i.e. unconscious use of syntax gives rise to the opportunity to perform automatic author identification based on words and style based features.

     In the paper, we have presented our approach of automatic authorship verification from a dataset containing cross-genre and cross-topic text snippets available in four different languages- Dutch, English, Greek and Spanish. In each of the problem sets, a group of known documents and one unknown document were provided. The goal of the present task is to predict if the author of an unknown document is same as that of the known document set by analyzing the similarity among the known documents and their differences (or similarities) to the unknown one. We have used 17

types of word based and style based features in total to find out the underlying similarities and differences of a set of documents. The Random Forest classifier that use a decision tree based approach and available with Weka[1] tool was employed in our system to choose the important features as we were not sure about the importance of a specific feature out of the main features viz. punctuation, sentence length, vocabulary, N-gram, Parts-of-Speech (POS).

The rest of the paper is as follows. Section 2 consists of a brief discussion on the related work available till date. Section 3 describes the feature selection for implementing the Random Forest algorithm whereas Section 4 gives the details of the system architecture. In Section 5, the experiments with detailed analysis of results are presented. Finally, the conclusions and future scopes of the experiments are presented in Section 6.

## 2    Related Studies

It is observed that the two main factors that characterize a text are its content and style, and both can be used as means of authorship categorization. Generally, there are two kinds of problems in author identification, the first one being comparatively easier than the second one. In the first kind, a finite set of documents with known authors is given and the task is to comment on the authorship of the unknown documents. In the second kind of task, a set of documents of a particular author is given along with an unknown document and the task is to predict whether the unknown document is written by the same author or not.

In author identification research, different aspects influence the performance of the author classification task. Some of the aspects such as the language of texts used, length of the text snippets, number of authors and texts, types of features etc. are considered as important for the classification task. It is observed that the number of features is most often varied to determine the influence of certain types of features. Corney et al. indicates that the most successful features are the function words and character n-grams whereas McCombe performed the tests using word unigrams as classification feature, for which the results were promising [2][4]. But, no method was successful in classification based on word bigrams, which seems contradictory because word bigrams capture more information about the sentence structure used by the author. On the other hand, Hirst and Feiguina used tag bigrams to discriminate between the works of two authors with three experiments [5].

So far, all the researches we described used only the English data sets. In contrast, author identification tasks were also performed on messages of other languages. In order to identify authors of Greek texts published in a newspaper, Stamatatos et al. used a combination of lexical measures and style markers and achieved an accuracy of 87% in texts classification [6]. It has to be mentioned that all of the previous researches discussed have one thing in common: they all use less than 10 authors.

---

1  http://weka.wikispaces.com/Use+WEKA+in+your+Java+code

Houvardas and Stamatatos used a data set consisting of a training set containing 2500 texts from 50 authors, with features like the most frequently occurring character n-grams of variable length (3-grams, 4-grams and 5-grams) and achieved an accuracy of 73.08% [7].

In literary authorship, the stylometric features are commonly used, for example- Letter frequencies, N-gram frequencies, Function word usage, Vocabulary richness, Lexical richness, Distribution of syllables per word, Word frequencies, Hapax legomena, Hapax dislegomena, Word length distribution, Word collocations, Sentence length, Preferred word positions, Prepositional phrase structure, Distribution parts of speech, Phrasal composition grammar etc. [1][2][3][4][8][9]. The fact is that there is no such consensus on which stylometric features are applied to achieve the best results for authorship identification.

In case of previous year's PAN and especially in the author identification task, Khonji and Iraqi used lots of features with parameter tuning in a GI framework to obtain a final score of 0.490 as product of AUC and c@1 [10].

## 3     Feature and Classifier Selection

It is found that several possible features from different categories can be used for author identification task. Thus, the number and types of features are often varied in author identification research to determine the influence of certain types of features. As this year's task includes cross-genre and cross-topic datasets, the features based on word-sense won't be much helpful compared to word structure based features. Therefore, we have used a combination of word and style based features in our present system.

- **Total Punctuation Count:** This feature counts the number of total punctuation symbols used in a text, normalized by the word count in that text.
  *For example:*
  *"We are participating in PAN, a part of CLEF 2015. Our main focus will be on Authorship Verification task- a subset of Author Identification."*
  *Here, this particular feature value is: (number of punctuation/number of words) = 4/24 = 0.16666*
  *We will calculate differences between unknown text and each known text of a problem set and take the average of those differences.*

- **Specific Punctuation Ratio:** This is the ratio of the total number of specific punctuation symbols like comma (,), semicolon (;), question-mark (?), exclamation-mark (!), stop (.), slash (/), dash (-), colon (:) etc. to the total punctuation count.
  *For example:*
  *"We are participating in PAN, a part of CLEF 2015. Our main focus will be on Authorship Verification task- a subset of Author Identification."*

*Here, the feature values will be:*
*StopCount= (number of stop/total number of punctuation) = 2/4 = 0.5*
*CommaCount= (number of stop/total number of punctuation) = 1/4 = 0.25*
*DashCount= (number of stop/total number of punctuation) = 1/4 = 0.25*
*ColonCount= (number of stop/total number of punctuation) = 0/4 = 0.0*
*ExclamationCount= (number of stop/total number of punctuation) = 0/4 = 0.0*
*QuestionCount= (number of stop/total number of punctuation) = 0/4 = 0.0*
*SlashCount= (number of stop/total number of punctuation) = 0/4 = 0.0*
*SemicolonCount= (number of stop/total number of punctuation) = 0/4 = 0.0*
*We will calculate differences between unknown text and each known text of a problem set and take the average of those differences in a similar manner like the previous one.*

- **Long-sentence/ Short-sentence Ratio:** Ratio of the long (length>12) or short (length<6) sentences to the total number of sentences is represented by this feature.
  *For example:*
  *"We are participating in PAN, a part of CLEF 2015. Our main focus will be on Authorship Verification task- a subset of Author Identification."*
  *Here, first sentence has length of 10 and second sentence has length of 14. So we have 0 short sentences and 1 long sentence.*
  *LongSentenceRatio=1/2=0.5*
  *ShortSentenceRatio=0/2=0.0*
  *Then again, we take the differences in a similar manner.*

- **Vocabulary Strength:** We tried to capture the vocabulary strength of an author by calculating the ratio of the unique words used to the total number of words used in a text snippet.
  *For example:*
  *Known-"We are participating in PAN, a part of CLEF 2015."*
  *Unknown- "We are participating mainly in an Authorship Verification task- a subset of Author Identification task."*
  *Here, in Known, total number of words= 10 and unique words= 10.*
  *In Unknown, total number of words= 15 and unique words= 13 (without stemming and stopword removal)*
  *So, VocabularyStrength in Known=10/10=1.0.*
  *VocabularyStrength in Unknown=13/15=0.8666.*
  *So, the feature value will be: 0.1333*

- **N-gram Difference:** This particular feature is very common for the task of authorship verification, where we tried to capture the n-gram (bigram and trigram for our system) overlaps among the known and unknown texts.
  *For example:*
  *Known-"We are participating in PAN, a part of CLEF 2015."*
  *Unknown- "We are participating mainly in Authorship Verification task- a*

*subset of Author Identification."*
*Here, some of the bigrams are- 'We are', 'are participating', 'participating in' etc. and some trigrams are- 'We are participating', 'are participating in' etc. (without stemming and stopword removal)*
*BigramOverlap=2 ['We are', 'are participating']*
*TrigramOverlap=1 ['We are participating'], for this particular example.*

- **POS Frequency:** In this feature, we try to capture the tendency of an author to use one or two particular types of POS that appear more frequently than the others, if there is any. So, we calculate the frequencies of each POS tag from texts and compare the known and unknown texts based on that.
  *For example:*
  *Known-"We are participating in PAN, a part of CLEF 2015."*
  *Unknown- "We are participating mainly in Authorship Verification task- a subset of Author Identification."*
  *Here, the POS tags are: We/Pronoun are/AuxVerb participating/Verb in/Preposition PAN/ProperNoun, a/Det part/Noun of/Preposition CLEF/ProperNoun 2015. Our/Pronoun main/Adjective focus/Noun will/AuxVerb be/Verb on/Preposition Authorship/Noun Verification/Noun task/Noun- a/Det subset/Noun of/Preposition Author/Noun Identification/Noun.*
  *Feature value: 6*

- **POS Sequence Frequency:** The feature is similar to the previous feature, but, here we try to find out the similarities based on a particular sequence (in a span of two consecutive POS tags) that the author uses for a considerable number of times.
  *In the above example, we have sequence overlaps: 'AuxVerb Verb', 'Verb Preposition', 'Det Noun', and 'Noun Preposition'. So, feature value is: 4*

- **Starting POS Frequency:** We try to list the POS tags that the author uses in the beginning of sentences according to their frequency and then compare them among the known and unknown documents to find a lexical pattern. For example, a particular author might have the tendency to start sentences with auxiliary verbs (example) or prepositions (in, for) unknowingly for a considerable number of sentences in the corpus. The feature also indicates the writing style of the author.
  *In the above example, both known and unknown text starts with the POS tag Pronoun. So, feature value is 1.*

As it is impossible to decide manually which of these features are most important or relevant to our problem structure, we decided to go for Decision Tree based classifier. Such classifiers are fast to train and easy to evaluate and interrupt and moreover non-parametric and for the very reason, we don't have to worry about the outliers or whether the data is linearly separable or not. The main disadvantage is that they easily overfit, but that's where ensemble methods like Random Forest come in. The main advantages of such approaches are:
- Almost always have lower classification error and better F-scores than decision trees.
- Almost always perform as well as or better than SVMs, but are easier to understand.
- Deal really well with uneven data sets that have missing variables.
- Give a good idea of which features in the data set are most important.
- Generally train faster than SVMs.

## 4　System Architecture

Figure 1 and 2 below illustrates the basic step-by-step architecture of our training and testing software, respectively. We were given datasets in four different languages- Dutch, Spanish, English and Greek, which contain numbers of known and unknown text samples by several authors. In a single author subset, we had one or more documents that are known to be written by that author and an unknown one for which the authorship is not known. The task is to predict whether that particular unknown document is written by the same author of that author subset or not. A .json file contains the language and the problem titles of a particular language dataset. For training sets, we were given a .txt file with the answers to those subset problems, i.e. whether or not those unknown files of a subset problem are written by the same author of other files in that subset.

At first, we read the contents from the text files and convert them all to the lowercase for an efficient parsing. Using the Stanford CoreNLP[2] parser, we tagged the contents of files to obtain the root words and POS tag for each word. Next, we counted the frequencies of each root word, all POS tag, POS tag-sequence and the starting POS tags from the tagged output and calculated the bigram and trigram frequencies using root-words as well. Also, the vocabulary strength of an author, i.e. the ratio of number of different types of root words to number of total words occurred is another feature that we considered in our approach. The long and short sentence ratio and punctuation counts are calculated from the plain text, i.e. raw file contents. We used the API of Weka 3.7.7.5 tool to employ a Random Forest classifier with the .arff file written using the features that we extracted from the training dataset. Weka is an open source data mining tool. It presents a collection of machine learning algorithms for data mining tasks.

---

[2] http://www-nlp.stanford.edu/software/corenlp.shtml

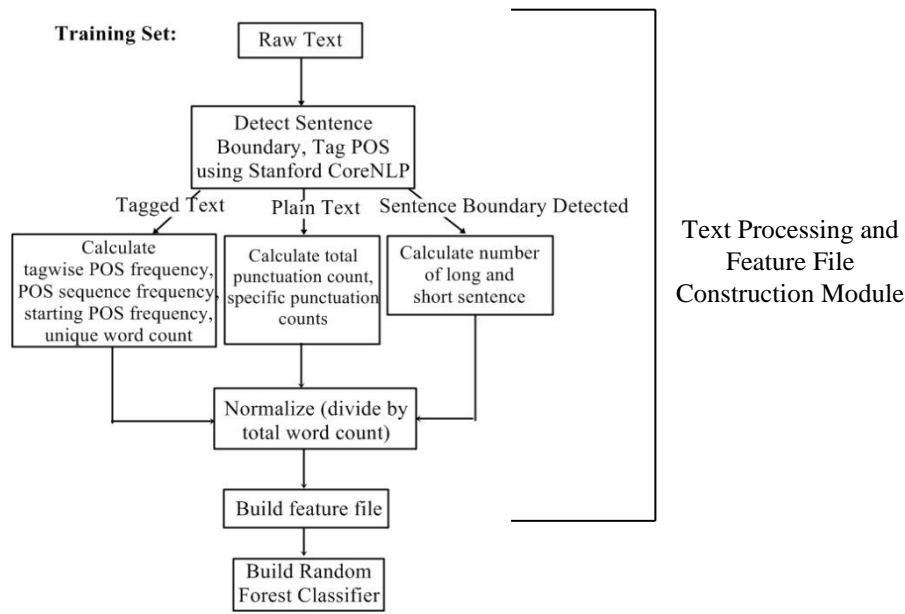

Fig. 1

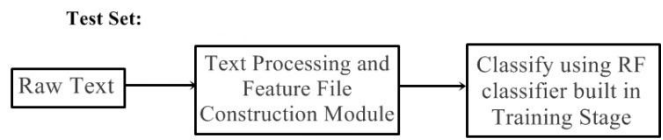

Fig. 2

In case of testing the dataset, after extracting, calculating features and writing in .arff file, we used the same classifier to predict the answer class for unknown documents in test dataset.

## 5  Experiment and Analysis

Table 1 shows the detailed results of our system in PAN 15 authorship verification task as provided by PAN organizing committee. Our present approach achieved the third best result in Dutch language with an AUC score of 0.75874 and C@1 score of 0.68283. We observe that our system didn't perform well in other languages like English, Greek and Spanish and obtained AUC scores 0.60174, 0.6126, 0.6096 and C@1 scores 0.57732, 0.5824 and 0.5768, respectively. One possible reason behind this fluctuation in result might be because of the variable number and size of the 'known'-documents. Cross-topic and cross-genre texts appear to be more common in English and Spanish, which produced a visible decline in the performance. To deal with multi-genre, we trained our system to analyze multiple corpuses/genre-specific training data and build the trainer as a whole. However, we had to modify our code as PAN committee decided to append time stamps automatically to the output folder and therefore the merging of the training sets was not possible.

| Language | AUC | C@1 | FinalScore | Time |
|---|---|---|---|---|
| **Dutch** | **0.75874** | **0.68283** | **0.51809** | 02:32:48 |
| English | 0.60174 | 0.57732 | 0.34749 | 15:19:13 |
| Greek | 0.6126 | 0.5824 | 0.35678 | 06:22:48 |
| Spanish | 0.6096 | 0.5768 | 0.35162 | 10:36:31 |

Table: Results on the Test Sets

## 6   Conclusion and Future Scope

In this paper, we have presented our approach to build software for automatic authorship verification using methods of Text Analytics. It uses same features for all the language datasets and a Random Forest classifier to classify the unknown documents based on the features extracted. In the recent years, the practical applications for authorship attribution have grown in areas such as intelligence (linking intercepted messages to each other and to known terrorists), criminal law, civil law and computer security (tracking authors of computer virus source code). This activity is part of computer science for identifying technologies, including biometrics, cryptographic signatures, intrusion detection systems and others.

In our future work, the accuracy of system can be improved by including some language specific features. . On the other hand, the features like the average paragraph length, average word length along with a Deep Learning classifier might produce interesting hike in our system score.

### References

1. Argamon, S., Koppel, M., Pennebaker, J. W., and Schler, J.: Automatically profiling the author of an anonymous text. Communications of the ACM, 52(2), 119-123 (2009).

2. Koppel, M., Schler, J., and Argamon, S.: Computational methods in authorship attribution. Journal of the American Society for information Science and Technology, 60(1), 9-26 (2009).

3. Corney, M.: Analysing e-mail text authorship for forensic purposes. Master's thesis, Queensland University of Technology, Brisbane Australia, 2003.

4. McCombe, N.: Methods of author identification. Master's thesis, Trinity College, Dublin Ireland, 2002.

5. Hirst, G. and Feiguina, O.: Bigrams of syntactic labels for authorship discrimination of short texts. Literary and Linguistic Computing, 22(4), 2007.

6. Stamatatos, E., Fakotakis, N. and Kokkinakis, G.: Computer-based authorship attribution without lexical measures. In Computers and the Humanities, 193–214, 2001.

7. Houvardas, J. and Stamatatos, E.: N-gram feature selection for authorship identification. In AIMSA, pages 77–86, 2006.

8. Patra, B. G., Banerjee, S., Das, D. and Bandyopadhyay, S.: Feeling may separate Two Authors: Incorporating Sentiment in Authorship Identification Task. In 10th International Conference on Natural Language Processing (ICON 2013), Delhi, India, 121-126.

9. Patra, B. G., Banerjee, S., Das, D., Saikh, T. and Bandyopadhyay, S.: Automatic Author Profiling Based on Linguistic and Stylistic Features-Notebook for PAN at CLEF 2013. In 9th evaluation lab on uncovering plagiarism, authorship, and social software misuse (PAN 2013) with CLEF-2013.

10. Khonji, M. and Iraqi, Y.: A Slightly-modified GI-based Author-verifier with Lots of Features (ASGALF). In proceedings of PAN at CLEF, 2014.